\documentclass[11pt]{article}

\usepackage[preprint]{requirements/acl}

\usepackage{times}
\usepackage{latexsym}
\usepackage[T1]{fontenc}
\usepackage[utf8]{inputenc}
\usepackage{microtype}
\usepackage{inconsolata}
\usepackage{graphicx}
\usepackage{algorithm}
\usepackage{algpseudocode}
\usepackage{multirow}
\usepackage{booktabs}
\usepackage{xspace}
\usepackage{lipsum}
\usepackage[most]{tcolorbox}
\usepackage{amsmath, amsfonts, amssymb}

\newcommand{\btree}{\textsc{BODHI-Tree}\xspace}
\newcommand{\CPE}{$\textrm{CPE}$\xspace}
\newcommand{\DCPE}{$\Delta\textrm{CPE}$\xspace}
\newcommand{\passk}{$\textrm{pass}@k$\xspace}
\definecolor{boxcol}{HTML}{DBECFF}

\title{{BODHI}: Do LLMs Branch Out and Discover Heterogeneous Inferences?}

\author{
  \textbf{Soumadeep Saha\textsuperscript{$\dagger$}},
  \textbf{Krish Sharma\textsuperscript{$\dagger$}},
  \textbf{Akshay Chaturvedi\textsuperscript{*}},\\
  \textbf{Nicholas Asher\textsuperscript{$\dagger$}} \\\\
  \textsuperscript{$\dagger$}ANITI, University de Toulouse; \textsuperscript{*}LINAGORA Labs \\
  \small{\textbf{Correspondence:} \href{mailto:soumadeep.saha97@gmail.com}{soumadeep.saha97@gmail.com}}
}

\begin{document}

\nolinenumbers
\maketitle

\begin{abstract}
    Although reinforcement learning with verifiable rewards (RLVR) has improved
    the reasoning performance of large language models (LLMs), whether it
    expands reasoning capabilities or primarily improves sampling efficiency
    remains debated. In this paper, we investigate the nature of test-time
    exploration in RLVR-trained LLMs by employing controlled maze-solving
    experiments and \btree{}s, a reusable representation that groups
    mathematical reasoning segments based on semantic equivalence. These probes
    help delineate preferences between stylistic variations and semantically
    distinct continuations. We show that RLVR-trained models exhibit lower
    candidate preference entropy between semantically distinct continuations
    than their distilled counterparts, and complementary sampling experiments
    show lower empirical semantic entropy and fewer observed branches under
    equal sampling budgets. Thus, policy concentration extends beyond stylistic
    choices and reduces observed semantic diversity. This reduction coexists
    with improved constraint adherence and backtracking in mazes, and improved
    recovery from distracting steps in mathematical reasoning. We conjecture
    that improved trace validity drives RLVR's sampling-efficiency gains, while
    policy concentration reduces rollout diversity. Our findings identify a
    central challenge for RLVR: preserving diverse reasoning trajectories
    alongside improved trace validity.
\end{abstract}

\section{Introduction}
\label{sec:intro}

\begin{figure*}
    \begin{center}
        \includegraphics[width=\textwidth]{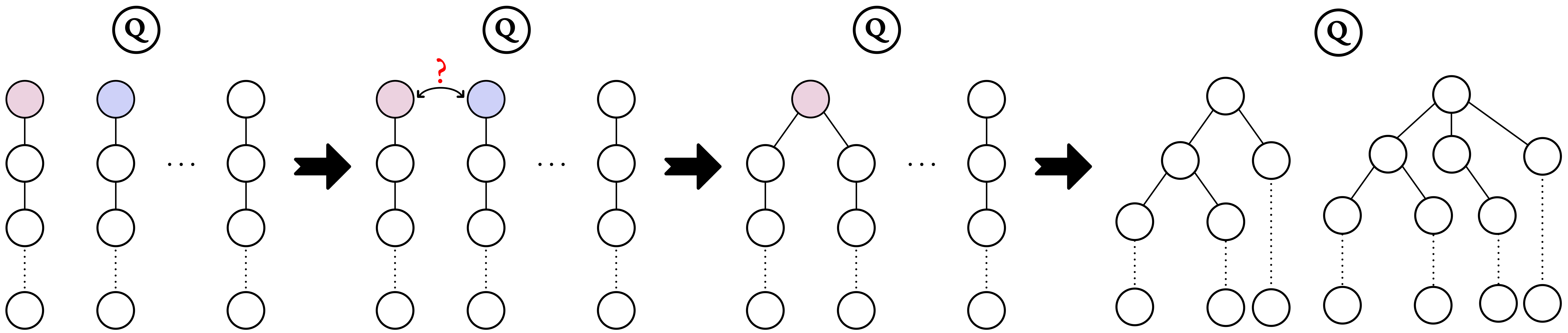}
    \end{center}
    \caption{
        \textbf{Construction of mathematical reasoning trees.}
        We first segment reasoning traces into a chain of reasoning nodes,
        followed by merging said nodes based on semantic similarity. Further
        details are provided in Appendix \ref{sec:appendix-aimetree}.
    }
    \label{fig:aimetree}
\end{figure*}
\footnote{All associated artifacts can be found at \href{https://www.example.com}{\texttt{anonymized}}.}%
Several recent papers have pointed to a curious problem in \emph{reinforcement
learning with verifiable rewards} (RLVR)-trained large language models (LLMs).
These models do not seem to ``discover'' novel abilities beyond the base model,
are less exploratory, and are reliant on a minority of high-entropy tokens to
drive reasoning diversity \citep{chen-etal-2026-dra, saha2025kismath,
wang2025beyond, yue2025does}. An exploratory model with the ability to access a
large number of \emph{valid, semantically distinct} continuations is evidently
desirable---it allows for the realisation of novel solutions, increased
diversity, and even performance scaling with techniques such as self-consistency
\citep{selfconsistency}, tree-of-thought \citep{yao2023tree}, etc. However, it
is also clear that \emph{not all exploration is desirable}. For instance, an LLM
with the capability to produce a large set of \emph{valid continuations}---%
correct mathematical reasoning traces---which differ in superficial ways such as
the naming of variables, order of commutative operations, or minor syntactic
variations in its natural language (NL) framing, does not constitute meaningful
exploration as it fails to traverse fundamentally distinct reasoning pathways
or yield novel solution strategies.

This brings us to the core tenet motivating this study: \emph{do LLMs Branch Out
and Discover Heterogeneous Inferences}, or is the exploration largely
decorative? To this end, we pose three \textbf{research questions} in this study:
\par\noindent $\diamond$ \textbf{RQ1}: Does RLVR change preferences between syntactic variants versus semantically distinct continuations?
\par\noindent $\diamond$ \textbf{RQ2}: Do these changes correspond to reduced semantic diversity in sampled continuations?
\par\noindent $\diamond$ \textbf{RQ3}: What are the behavioral implications of this policy shift \emph{vis-a-vis} performance?

Surprisingly, despite many recent papers making fundamental improvements to our
understanding of the dynamics of RL training, especially pertaining to
preventing policy collapse, encouraging exploration, etc. \citep{yu2026dapo,
drgrpo, treepo, treerpo, zhao-etal-2026-reinforced}, the fundamental question of
\textbf{how RLVR changes preferences} among semantically distinct reasoning
continuations remains under-explored. We investigate exploration in LLMs through
two modalities: \textbf{mazes}, where the notion of exploration is rather
straightforward, and \textbf{mathematical reasoning}, where it is much less
straightforward due to the complexities of the mathematical and linguistic
expressions used.

Our work studies the effect of post-training on policy preferences between
semantically distinct continuations, and not entropy collapse or its mechanism.
\textbf{The core contributions of our work are as follows:} \textbf{(i)} in
order to investigate exploration in \emph{mathematical reasoning}, \textbf{we
concretize the notion of exploration} by collapsing a large number of alternate
traces corresponding to a problem into a tree structure (\btree{}, see Fig.
\ref{fig:aimetree}), wherein each node represents several semantically identical
reasoning segments and its distinct children represent conceptually different
continuations, thus allowing for fine-grained analysis of the nature of
exploration; \textbf{(ii)} Using our \btree{}s and mazes, we demonstrate that
the policy resulting from \textbf{RL-training shows semantic concentration} and
not just syntactic preference calcification, a finding further validated by
on-policy sampling showing lower empirical semantic entropy and fewer observed
branches under equal sampling budgets; and \textbf{(iii)} we analyze different
properties associated with exploration such as local validity and backtracking,
and find that reduced exploration does not imply poorer recovery, and RLVR
improves constraint adherence and backtracking. This leads us to proffer that
improved trace validity could be a simple explanation for RLVR's sampling
efficiency gains, while policy concentration reduces trajectory diversity.
Additionally, in the course of this study we created and open-sourced several
fully-reproducible intermediate post-training checkpoints (in addition to
datasets and code), to aid future work into RLVR dynamics.

\section{Background}
\label{sec:background}

\emph{RLVR} reframes next-token prediction as a Markov Decision Process,
learning a policy $\pi_{\theta}$ for predicting next tokens $o_i$, that leads to
trajectories of states
$[\textrm{Input}] \rightarrow [\textrm{Input}, o_1] \rightarrow \ldots \rightarrow [\textrm{Input}, o_1, \ldots o_T]$,
that maximize expected reward from deterministic verifiers. This paradigm has
significantly improved overall performance on several reasoning and planning
benchmarks, and has pushed the frontiers of capabilities such as integration of
tools and \emph{agentic} applications. It also enables an ancillary technique,
called \emph{Long-CoT distillation} or simply \emph{distillation}
\citep{deepseekr1, shao2024deepseekmath}, wherein an LLM is fine-tuned (SFT) on
traces generated from a more capable RL-trained ``teacher'' LLM.

Given a state $s_t$, the set of continuations---every finite string with $s_t$
as a prefix---can be segmented into various classes which are
``\emph{verifier equivalent}''. However, since the verifier often only evaluates
the last few tokens of the terminal state \citep{shao2024deepseekmath,
deepseekr1, yu2026dapo, drgrpo}, verifier equivalence does not cleanly map to
semantic equivalence. For instance, a specious continuation which happens to
present the correct answer has the same utility as a semantically correct
continuation, and there is no a priori guarantee that verifier-equivalent but
semantically diverse continuations have an equal propensity of realization
\citep{anschel-etal-2025-group}. This serves as the core object of analysis in
our study: the \emph{accessibility of different verifier-equivalent trajectories}%
\footnote{i.e., the probabilities of different (finite) continuations that are verifier-equivalent.}
at \textbf{test time}, a property we refer to as \textbf{exploration}. This is
different from \emph{inter-trace exploration} \citep{jiang-etal-2025-makes}, and
asks the counterfactual question: \emph{what other continuations were likely?}

Test-time exploration elicits the ``reasoning capability'' boundary
\citep{yue2025does}, since, given enough samples, a model capable of accessing
diverse trajectories will arrive at novel solution approaches or techniques.
Thus, a litmus test for the efficacy of RL-training is its ability to explore.
Additionally, exploration can amplify performance with the aid of accessory
ensembling techniques such as \emph{tree-of-thought} \citep{yao2023tree}. With
the proliferation of RL-trained LLMs, it is therefore critical to understand the
\emph{impact of the reward signal on the model's generative policy}, especially
since the assumption that RLVR consistently incentivizes semantically diverse
and meaningful exploration has been widely challenged
\citep{saha2025kismath, wang2025beyond, yue2025does, anschel-etal-2025-group}.

\citet{yue2025does} claimed that RL-trained LLMs have degraded \passk scaling
and do not discover fundamentally novel reasoning pathways beyond the base LLM.
\citet{saha2025kismath} posited that RL-trained models are ``over-confident''
compared to distilled models: they have a lower entropy policy which is
correlated with a drop in performance. Similarly, \citet{wang2025beyond}
demonstrated that the RL-policy is reliant on a minority ($\sim 20\%$) of
high-entropy ``forking tokens'' to drive meaningful exploration at test time
\citep{cheng2025reasoningexp}, and \citet{jang-etal-2026-bad} attributed this
collapse to ``over-confidence at a small number of structurally critical
decision points''. \citet{yuan2026undvercol}, however, oppose the view that
decline of high-k \passk performance is indicative of a decline in reasoning
diversity. They liken RLVR exploration collapse to ``overtraining'' and proffer
re-weighting policy updates to favor low-success rollout groups.
\citet{cai2026currrein} make a similar diagnosis, and advance ``boundary-aware''
curriculum learning to enhance the frontier of reasoning capabilities.
\citet{huang-etal-2026-semantic} argue that token-level statistics do not
reflect how reasoning progresses over multi-token semantic structures. This
debate underscores a critical gap: \emph{existing performance-based and
token-level metrics do not indicate if entropy collapse implies a lack of
reasoning diversity}.

The collapse of policy entropy during RL-training has received significant
attention, and several recent papers propose modifications to the standard
RL-training regimen. Monte Carlo Tree Search-based algorithms
\citep{treepo, treerpo, fr3e} that sample rollouts during training in a
tree-like structure and assign segment-level credits have been proposed. Other
techniques explicitly include cues to enhance exploration based on properties of
the rollout group \citep{chen-etal-2026-dra}. Techniques such as those based on
occurrence frequency \citep{anschel-etal-2025-group}, embedding similarity
\citep{zhao-etal-2026-reinforced}, feedback from auxiliary models
\citep{mishra-etal-2026-sd, hu-etal-2026-rewarding}, and many others 
\citep{li-li-2026-lad, he-etal-2026-vane, cai2026currrein, hu2026diversityincen,
jang-etal-2026-bad, yuan2026undvercol, huang-etal-2026-semantic, chen2026passk}
have been proposed, with varying degrees of success
\citep{zhao-etal-2026-reinforced, hao-etal-2026-rethinking}.
\citet{hao-etal-2026-rethinking} study factors influencing entropy collapse
during training and highlight that recent approaches rely on heuristic
adjustments to some of these factors, thus limiting their efficacy.

In related studies, \citet{jin-etal-2026-revisiting} studied the relationship
between model performance and training entropy, finding a negative but
task-dependent correlation. \citet{wen2026reinforcement} used an
\emph{LLM-as-a-judge} framework to evaluate intermediate trace correctness
(CoT-\passk) alongside the answer, demonstrating that RLVR-trained models have
enhanced trace validity compared to the base model. The results for CoT-\passk
scaling in mathematical reasoning tasks are mixed, and the authors speculate
that the \emph{``distilled LLM may already master major reasoning capabilities
that can be learned with RLVR''}, which further underscores the need to isolate
the effect of RLVR post-training on test-time exploration.
\citet{jiang-etal-2025-makes} introduce LCoT2Tree, which parses Long CoT traces
into trees by mapping segments of reasoning traces to an extracted high-level
summary. By classifying segment functions (e.g., verification, exploration,
backtracking), they identify structural error patterns such as over-branching.

Our work, instead of focusing on token-level metrics or performance alone,
attempts to measure \textbf{how RLVR changes preferences among semantically
distinct verifier-equivalent reasoning continuations}, and tries to ascertain if
\emph{the entropy collapse is simply an artifact of stronger preferences for
syntax and style} or if the model also \textbf{has calcified preferences when it
comes to semantically diverse continuations}. Our contribution is \textbf{not}
the observation that post-training concentrates policies.

\section{Methodology: Measuring Exploration}
\label{sec:method}

In this section, we outline the construction of the \emph{probing datasets} that
enable us to distinguish differences between the policies of various LLMs with
regard to exploration. While NL \textbf{mathematical reasoning} directly
measures an LLM's logical capabilities, it is inherently noisy; the vast
linguistic state space allows models to generate superficial restatements or
varying syntactic formulations that are challenging to verify and can mimic
diversity without representing genuine inferential branching. Thus, we
additionally employ \textbf{maze solving}, to isolate the routing and planning
capabilities and completely eliminate effects of superficial restatements. We
report results with $2$-sample Kolmogorov-Smirnov test statistics ($D_{KS}$),
$p$ values, and bootstrapped 95\% CIs (in square brackets).

\subsection{Maze Exploration}

\begin{figure*}[h]
    \begin{center}
        \includegraphics[width=\textwidth]{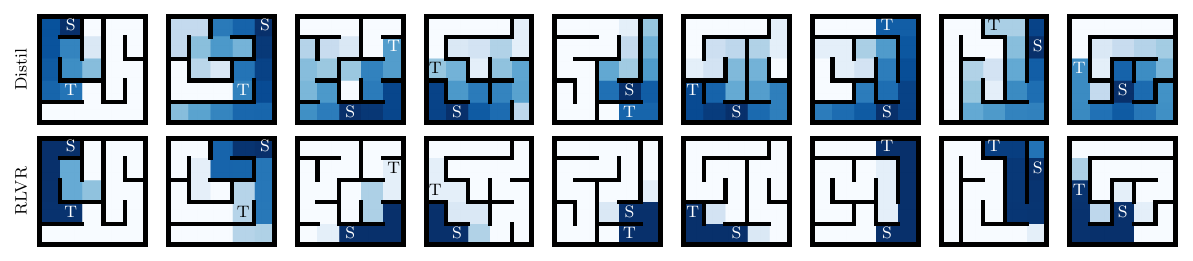}
    \end{center}
    \caption{
        \textbf{RL-trained models are less exploratory.} We plot how often a
        node is visited by the distilled and RLVR-trained \texttt{Qwen3-8B}.
    }
    \label{fig:mazeaccess}
\end{figure*}

As a probe, maze solving serves as a noise-free diagnostic setting in which
trace properties are directly identifiable, while preserving the fundamental
structure of step-by-step problem-solving---the model must sequentially execute
valid, discrete actions to reach a goal state. Since the action space is
strictly constrained, and local trace validity (e.g., avoiding walls) and
verifier-equivalent trace families are trivial to formalize, mazes provide an
ideal, controlled environment to characterize the nature of policy shift arising
from RL-based post-training.

In this study, we employed the 
\href{https://understanding-search.github.io/maze-dataset/maze_dataset.html}{Maze Dataset}
\citep{maze-dataset} library to instantiate various deterministic graph
traversal tasks, constraining the model's action space to four directional
moves: $\{\texttt{<left>}, \texttt{<right>}, \texttt{<up>}, \texttt{<down>}\}$.

\subsection{Exploration in Mathematical Reasoning}

\emph{Ideally}, to study exploration in NL mathematical reasoning, we would
start from a shared reasoning state and construct a tree whose edges represent
atomic, semantically distinct inferential actions to distinct reasoning states,
analogous to the maze setting. In practice, this construction faces two
challenges: the continuation space is vast, and semantic equivalence is
difficult to establish across NL reasoning traces. Moreover, broader sampling
reveals more of the semantic continuation space but increases the cost of
identifying equivalent continuations: pairwise comparisons among $n$ samples
scale as $\mathcal{O}(n^2)$. A practical design must therefore balance sampling
coverage against the computational cost of reliable semantic-equivalence
judgments. 

Our approach prioritizes the \emph{quality of semantic-equivalence judgments}
while amortizing their computational cost through a reusable \btree
representation. Given a question $Q$, we first sample reasoning traces from a
diverse set of frontier LLMs and retain those whose final answers match the
ground truth, yielding responses $A^{(1)}, A^{(2)}, \ldots, A^{(n)}$.%
\footnote{$n \approx 90$ per question, total API budget: \texttt{USD 3000}.}
We then use an LLM to segment each response into individual \emph{reasoning
steps} $a^{(i)}_j$, such that
$A^{(i)} = \textrm{concat}([a^{(i)}_1, a^{(i)}_2, \ldots, a^{(i)}_{n_i}])$.

Using these segments and an LLM-based semantic matching function, we construct a
prefix tree $T = (V, E)$, called a \btree (Fig.~\ref{fig:aimetree},
Algorithm~\ref{algo:aimetree}), where each node $v = \{a^{(i)}_{\alpha}\} \in V$
represents a set of semantically equivalent mathematical statements, and an edge
$u \leadsto v \in E$ if every statement in $v$ comes after some statement in $u$
in some reasoning trace $A^{(i)}$. The construction aligns traces by merging
segments judged semantically equivalent, with containment-based matching to
accommodate differences in granularity. Root-to-leaf paths therefore represent
alternate solution traces for $Q$, while branching points identify candidate
divergences in reasoning. This structure provides a \textbf{shared set of
contexts and alternative continuations} for comparing model preferences
head-to-head, including \emph{preferences between syntactic variants and
semantically distinct continuations}.

This shared probe set offers an important advantage over direct on-policy
sampling. Even when conditioned on the same prefix, different policies generally
produce different trace realizations; comparisons between their sampled outputs
therefore vary both the policy and the realized continuations. In contrast,
\btree probes enable \textbf{paired comparisons that hold the textual context
and candidate continuations fixed}, isolating changes in model preferences over
those candidates. Moreover, each \btree is constructed once and reused across
evaluated models, \emph{amortizing the cost of clustering across checkpoints}
and enabling large-scale comparisons without reconstructing trees for every
model.

These advantages come with a complementary limitation: fixed-candidate probes
delineate relative preferences, not the full on-policy continuation
distribution. We therefore supplement them with smaller-scale on-policy
experiments, sampling continuations directly from each model at selected
reasoning states and clustering them using their complete context. Together, the
two approaches combine \emph{controlled preference comparisons} with evidence
about the semantic diversity of continuations actually sampled from each policy.

$235$ questions from
\href{https://huggingface.co/datasets/gneubig/aime-1983-2024}{AIME}, alongside
$\sim20K$ responses and their corresponding trees constitute our probe dataset
\btree. Semantic similarity of two segments is gauged using
\texttt{GPT-oss-120b} in an \emph{LLM-as-a-judge} framework, and these
judgements are validated against other frontier models and checks by authors
(Cohen's $\kappa = 0.90$). Further details on the construction of \btree{}s,
like prompts, settings, validation of the \emph{LLM-as-a-judge} framework, 
error-analysis, etc., can be found in Appendix \ref{sec:appendix-aimetree}.

\subsection{Models}

Limited access to intermediate checkpoints (popular model vendors do not release
intermediate post-training checkpoints) and differences in post-training data
complicate comparisons across models. We therefore train most checkpoints
ourselves starting from publicly available pre-training checkpoints, and these
\textbf{intermediate post-training checkpoints are released}, enabling
comparisons beyond final model endpoints.

For \textbf{mathematical reasoning}, we evaluate two complementary designs.
\textbf{Matched Distil$\rightarrow$RLVR lineages} (\texttt{Qwen3-8B-Base} and
\texttt{gemma-3-12b-pt}; \citealp{qwen3,gemma3}) measure changes after adding
RLVR to the same distilled checkpoint. \textbf{Parallel distillation and Zero-RL
settings} (\texttt{Qwen3.5-9B-Base} and \texttt{Qwen2.5-32B-Base};
\citealp{qwen35,qwen25}) compare alternative post-training routes from the same
base model, covering \emph{widely used training configurations} rather than
sequential checkpoint evolution. Distillation uses $50$K samples from
\href{https://huggingface.co/datasets/open-r1/OpenThoughts-114k-math}{\texttt{OpenThoughts-114k-math}},
and RLVR uses \texttt{DAPO-Math-17k} \citep{yu2026dapo}; for the $32$B setting,
we adopt \texttt{DAPO-Qwen2.5-32B}. Matched lineages provide the primary
evidence for changes following RLVR, while parallel settings assess consistency
across training routes.

For \textbf{mazes}, we train matched SFT$\rightarrow$RLVR lineages using
\texttt{SmolLM3-3B-Base} \citep{smollm3} and \texttt{Qwen3-8B-Base}: SFT uses
$30$K oracle-generated solutions, followed by RLVR on $12$K mazes. Further
details appear in Appendix~\ref{sec:appendix-expdet}.

\subsection{Metrics}
\label{sec:metrics}

Since this work focuses on the policy distribution \emph{vis-a-vis} exploration,
a quantity of interest is \textbf{Candidate Preference Entropy} (\CPE). A branch
node from \btree (or the tree implicit in maze exploration) gives us a prefix
$a$, and at least two alternative completions $s^{(1)}$ and $s^{(2)}$. Given an
LLM $P_{\theta}$ we can compute:
\begin{equation*}
    S_{\theta}\big(s^{(i)} | a\big) = \frac{1}{|s^{(i)}|} \sum_{k = 1}^{|s^{(i)}|} \log\big(P_{\theta}(s^{(i)}_k | s^{(i)}_{<k},a)\big)\\
\end{equation*}
We normalize these scores to obtain
$P(s^{(i)} | a) = \tfrac{\exp(S_{\theta}(s^{(i)} | a))}{\sum_{i} \exp(S_{\theta}(s^{(i)} | a))}$,
and compute:
\begin{equation}
\begin{split}
    &\textrm{CPE}_{\texttt{model}}(s^{(1)}, s^{(2)}; a) = H_{\texttt{model}}^{\textrm{Branch}}(s^{(1)}, s^{(2)}) \\
    &=  -\sum_{i = 1,2} P(s^{(i)} | a) \log\big(P(s^{(i)} | a)\big)
\end{split}
\end{equation}
where $H_{\texttt{model}}^{\textrm{Branch}}(s^{(1)}, s^{(2)})$ is the
\textbf{conditional entropy} in the choice of $s^{(1)}$, $s^{(2)}$ (or \CPE),
given prefix $a$, and $s^{(i)}_k$ is the $k$th token of $s^{(i)}$. Further,
since we want to study differences resulting from post-training, we compute:
\begin{equation}
\begin{split}
    &\Delta\textrm{CPE}_{\texttt{model}}(s^{(1)}, s^{(2)}; a) = \Delta H_{\texttt{model}}^{\textrm{Branch}}(s^{(1)}, s^{(2)})\\
    &= H_{\texttt{model}}^{\textrm{Branch}}(s^{(1)}, s^{(2)}) - H_{\texttt{base}}^{\textrm{Branch}}(s^{(1)}, s^{(2)})
    \label{eq:bentropy}
\end{split}
\end{equation}
$\Delta H_{\texttt{model}}^{\textrm{Branch}}$ is the \textbf{difference in
conditional preference entropy} (or \DCPE) and measures changes in preferences
of branches arising from a certain post-training scheme. \CPE is a
\textbf{preference proxy} to isolate exploration differences, and \emph{does
not} account for the distribution over all semantic branches.

Computing $\Delta\textrm{CPE}(s^{(1)}$, $s^{(2)}; a)$ (Eq. \ref{eq:bentropy})
for $10$K tuples from \btree{}s reveals (see Table \ref{tab:entdropped}, Fig.
\ref{fig:fulldist}; Appendix \ref{sec:appendix-extra}) a
\textbf{statistically significant drop} in \DCPE for \textbf{RL-trained models}
compared to their \textbf{distilled} counterparts, i.e., they have lower
branching entropy, and this drop is \textbf{observed in nearly every sample}.
This indicates that RL-trained models have stronger preferences when choosing
between alternate completions at branch points, consistent with prior test-time
compute scaling results \citep{yue2025does}.

Further, we can select a branching node $\alpha$ (has at least two children),
and its two children $\beta$ and $\gamma$ from a \btree. Then, picking a
representative anchor trace $A$ from the node $\alpha$, two continuations $a, b$
from child $\beta$ that are semantically similar but have different syntax, and
a continuation $g$ from child $\gamma$ which has different semantics from
$a, b$, we can compute (as in Eq. \ref{eq:bentropy}):
$\Delta\textrm{CPE}_{\texttt{model}}^{\textrm{Syntax}} := \Delta\textrm{CPE}_{\texttt{model}}(a, b; A)$
and 
$\Delta\textrm{CPE}_{\texttt{model}}^{\textrm{Semantics}} := \Delta\textrm{CPE}_{\texttt{model}}(a, g; A)$.
These quantities can help pinpoint whether the entropy collapse in the RLVR
policy arises due to the model developing stronger preferences for trivial
mathematical representational choices (such as how variables are named or the
order of commutative operations) alone, or also encompasses lack of semantic
diversity such as the choice of the \textbf{mathematical approach} to use.

\section{Results: Semantic Concentration}
\label{sec:math-results}

\begin{figure*}[th]
    \begin{center}
        \includegraphics[width=\textwidth]{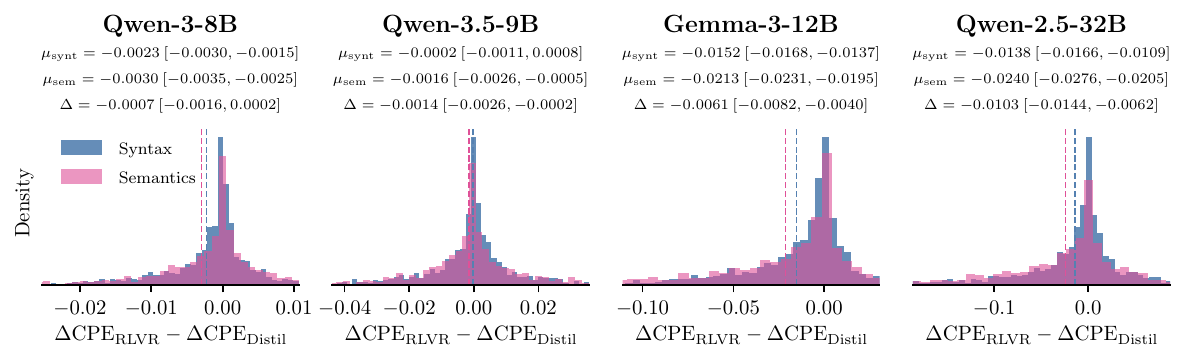}
    \end{center}
    \caption{
        \textbf{RLVR-induced policy concentration extends beyond stylistic
        variations:} across all four model families, RLVR reduces \CPE between
        semantically distinct continuations relative to distillation. The
        decrease is larger than that for syntactic alternatives in three of four
        families. The \emph{dashed} lines show the means of the distributions.
    }
    \label{fig:synt-semant}
\end{figure*}

\subsection{Semantic-Syntactic Comparison}
\label{subsec:style}

Our experiments show that \textbf{RLVR-induced policy concentration extends
beyond stylistic variations}. Fig. \ref{fig:synt-semant} plots the distribution
of the differences
$\Delta\text{CPE}_{\texttt{RLVR}}^{*} - \Delta\text{CPE}_{\texttt{Distil}}^{*}$
(see \S \ref{sec:metrics}) for $2.5$K sampled nodes from \btree{}s. Our
observations are: \textbf{(i) all RL-trained models in the study have a
statistically significant drop in $\Delta\textrm{CPE}^{\textrm{Semantics}}$
compared to their distilled counterparts}, \textbf{(ii)} there is a drop in
$\Delta\textrm{CPE}^{\textrm{Syntax}}$ for all RL-trained models compared to
their distilled counterparts (except for \texttt{Qwen3.5-9B}, where the drop is
not statistically significant), and \textbf{(iii)} for three out of four models
\textbf{the drop in $\Delta\textrm{CPE}^{\textrm{Semantics}}$ is larger}
($p < 0.02$ except for \texttt{Qwen3-8B}, $p = 0.11$). This suggests that the
policy distribution is \textbf{more collapsed when it comes to making choices
between semantically distinct continuations} and the policy collapse cannot be
explained by \textbf{just syntactic/stylistic} artifacts. The fact that
\emph{RL-trained models have a lower propensity for semantically meaningful
exploration} is important because this underlines the fact that RLVR does not
just eliminate spurious variations but also legitimate exploration.

\subsection{On-policy Branching}
\label{subsec:onpolicy}

\begin{table}
    \begin{center}
        \footnotesize{
        \begin{tabular}[c]{p{7em}|l l}
            \toprule
            \multirow{2}{*}{\textbf{Property}} & \multicolumn{2}{|c}{\textbf{RLVR - Distil}}                     \\
            \cmidrule{2-3}
                                               & \texttt{Qwen-3-8B}                   & \texttt{gemma-3-12b}     \\
            \midrule
            Largest cluster $P$                & 0.033  \scriptsize{[0.01, 0.06]}     & 0.070  \scriptsize{[0.04, 0.10]}     \\
            $H_{\text{emp}}$                   & -0.17  \scriptsize{[-0.27, -0.07]}   & -0.43  \scriptsize{[-0.55, -0.32]}   \\
            Branch Count                       & -2.25  \scriptsize{[-3.86, -0.69]}   & -7.97  \scriptsize{[-9.45, -6.40]}   \\
            $\text{CPE}_{\text{emp}}$          & -0.084 \scriptsize{[-0.10, -0.07]}   & -0.056 \scriptsize{[-0.08, -0.04]}   \\
            \bottomrule
        \end{tabular}}
    \end{center}
    \caption{
        \textbf{RLVR reduces on-policy sampled semantic diversity.} Differences
        between RLVR and distilled checkpoints in matched lineages after
        sampling $128$ continuations at each of the $200$ shared prefixes. 
        RLVR increases the frequency of the largest semantic cluster while
        reducing empirical semantic entropy ($H_{\text{emp}}$) and the number of
        observed branches. The sequence-level empirical \CPE measure also
        decreases.
    }
    \label{tab:onpolicy}
\end{table}

We directly measure semantic entropy collapse \emph{on-policy} for $200$ \btree
nodes by sampling $128$ continuations for each prefix, and clustering them based
on semantic similarity as assessed by the LLM judge (\texttt{GPT-oss-120b}).
The results \textbf{confirm} conclusions from \S\ref{subsec:style} (see Table
\ref{tab:onpolicy}): we find that the \textbf{empirically observed entropy}
($H_{\text{emp}}$) and number of observed branches (Branch Count)
\textbf{decrease}, whereas the probability mass in the largest cluster
\textbf{increases}. The on-policy \CPE analog\footnote{The sequences sampled
on-policy will be different between policies.}
($-\sum_{i} P(s^{(i)} | a) \log(P(s^{(i)} | a))$ for sequences $s^{(i)}$)
considering the full sampled distribution also shows a \textbf{decrease for the
RL-trained model}. This shows that RL-trained models have \textbf{reduced
observed semantic diversity under matched sampling budget}.

\section{Results: Maze Exploration}
\label{sec:maze-results}

\subsection{Where does the entropy go?}
\label{sec:maze-exp}

\emph{The exploration collapse in RLVR is also seen in mazes};
Fig.~\ref{fig:mazeaccess} plots the visitation frequency of each node in the
maze based on $1$K rollouts per maze by the RLVR and distillation policies, and
clearly shows this collapse. The entropy of the observed visitation distribution
drops from $H_{\text{Distil}} = 2.3380$ $[2.1815, 2.4899]$ to
$H_{\text{RLVR}} = 1.8138$ $[1.6477, 1.9755]$
($\Delta H = -0.5242$; $[-0.7005, -0.3435]$, $p < 0.0001$) for \texttt{Qwen3-8B},
and from $H_{\text{Distil}} = 1.9288$ $[1.7938, 2.0632]$ to 
$H_{\text{RLVR}} = 1.6092$ $[1.4344, 1.7756]$
($\Delta H = -0.3196$; $[-0.4252, -0.2085]$ $p < 0.0001$) for \texttt{SmolLM3-3B}.

\begin{figure}[h]
    \begin{center}
        \includegraphics[width=\linewidth]{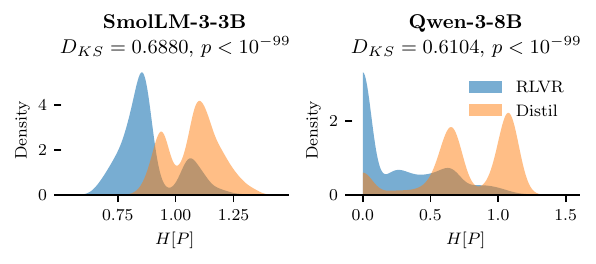}
    \end{center}
    \caption{
        \textbf{RL-trained models successfully learn the semantics of mazes.} 
        We plot the entropy of the models at hall nodes (degree 2) to see if
        different training modalities produce models which correctly model the
        lack of ambiguity at halls.
    }
    \label{fig:synt-semant-maze}
\end{figure}

A part of the observed entropy collapse could be explained by the model
assigning \textbf{less probability mass to invalid continuations}---an expected
property from well-behaved policies. To investigate this possibility, we choose
$5$K \textbf{``hall''} nodes (nodes with degree 2) along maze solution paths,
and since there is no ambiguity in the choice of next-token, a semantically
faithful policy should have lower entropy at these nodes. We find that (see Fig.
\ref{fig:synt-semant-maze}), the
\textbf{RL-trained models demonstrate statistically significant entropy collapse
in halls}, suggesting that the RLVR policy better delineates between different
verifier-equivalence classes. An equivalent notion for mathematical reasoning is
``local trace coherence'', i.e., the ability of a model to avoid logical errors
within sub-segments of traces, and there is preliminary evidence suggesting that
RLVR-trained models perform better in this regard \citep{samineni2025locarvity}.

\begin{table}[h]
    \begin{center}
        \begin{tabular}[c]{l|r|r}
            \toprule
            \multirow{2}{*}{\textbf{Decoding Policy}}
                                        & \multicolumn{2}{c}{\textbf{Training Method}}  \\
            \cmidrule{2-3}
                                        & Distil            & RLVR                      \\
            \midrule
            \texttt{Baseline}           & $7.6 \pm 1.2$   & $13.0 \pm 1.5$              \\
            \texttt{Legal Only}         & $62.6 \pm 2.2$  & $39.2 \pm 2.2$              \\
            \texttt{Avoid Short Path}   & $3.8 \pm 0.9$   & $4.6 \pm 0.9$               \\
            \bottomrule
        \end{tabular}
    \end{center}
    \caption{
        Causal interventions demonstrating increased trace validity and 
        policy concentration in RLVR.
    }
    \label{tab:mazepolicy}
\end{table}


A natural next question is: \emph{does RLVR maintain probability mass on valid
continuations}, or are \emph{valid, semantically distinct continuations also
suppressed?} To answer this, we design the following interventions on the
decoding strategy: \textbf{(i)} ``\texttt{Legal Only}'' where invalid next
tokens (moves that result in wall collisions) are masked out, and \textbf{(ii)}
``\texttt{Avoid Short Path}'' where the model is steered away from the shortest
path. The \texttt{Legal Only} intervention sets the probability of invalid
continuations to zero, and thus if a particular policy already had very low
likelihood of selecting these continuations, it will be affected less.
Similarly, the \texttt{Avoid Short Path} intervention sets the probability of
the optimal continuation to zero,%
\footnote{The intervention masks at most one move. An alternative path requiring
$3.4\pm0.3$ additional steps remains available, as is backtracking to the
shortest path after intervention.}
and thus should have a larger effect on a policy which has stronger trajectory
preferences. Table \ref{tab:mazepolicy} presents our results with the distilled
and RL-trained variants of \texttt{Qwen3-8B} (\texttt{T=1.0}) with $500$ mazes.

We see that the \texttt{Legal Only} intervention has only a $\sim3\times$
improvement on the RL-trained model, which is dwarfed by the $\sim8\times$
improvement on the distilled model. The distilled model with the
\texttt{Legal Only} constraint shows a $\sim60\%$ improvement over the
RL-trained model, and for the \texttt{Avoid Short Path} intervention performance
is reduced by $\sim65\%$ for the RL-trained model but only by $\sim50\%$ for the
distilled model. Taken together, this indicates that \textbf{RL-trained models
also have reduced exploration within valid verifier-equivalent distinct
continuations}: despite its higher baseline accuracy ($13\%$) and
de-prioritization of invalid moves, the RLVR model underperforms the distilled
model when invalid moves are masked, consistent with reduced trajectory
diversity. This policy ossification (other valid continuations are less likely
to be explored) occurs during training (see Fig. \ref{fig:evolution}, Appendix
\ref{sec:appendix-extra}, \citet{cui2025the}), and could be a potential
explanation for \emph{how RLVR achieves better sample efficiency
($\textrm{pass}@1$ performance)}: it might be learning environmental constraints
and suppressing trajectory diversity.

\subsection{Backtracking and Recovery}
\label{subsec:backtrack}

A potential consequence of semantic diversity collapse is the reduction in the
ability to backtrack. At test-time, a policy might reach a state from which
further progress is impossible, and valid continuation requires navigating out
of these ``dead-ends'' to a preceding state followed by exploration of alternate
trajectories. Thus we ask: \emph{is semantic entropy reduction associated with
reduced backtracking?}

We evaluate models at $2.5$K dead-end states (states from which the goal is only
reachable if the last move is reversed, e.g., \texttt{<up>} followed by
\texttt{<down>}) with prefix-steering, and measure the probability of the
backtracking move ($p_b$). Our results indicate that RLVR training actually
\emph{enhances backtracking capabilities} compared to distillation. For
\texttt{SmolLM3-3B}, the average backtracking probability increases from
$\mathbb{E}[p_b] = 0.0647 \pm 0.0008$ (Distil) to $0.1687 \pm 0.0035$ (RLVR)
($\Delta = 0.1040$; $[0.0978, 0.1101]$, $p~<~0.0001$), and for \texttt{Qwen3-8B}
it increases from $0.0068 \pm 0.0010$ (Distil) to $0.1537 \pm 0.0062$ (RLVR)
($\Delta = 0.1469$; $[0.1348, 0.1589]$, $p~<~0.0001$). Furthermore, RLVR reduces
absolute failures ($p_b = 0$), with zero-probability instances dropping from
$2422 \rightarrow 1824$ for \texttt{Qwen-3-8B} and $476 \rightarrow 357$ for
\texttt{SmolLM-3-3B}. The distribution is plotted in Fig. \ref{fig:btrack} in
Appendix \ref{sec:appendix-extra}.

Analyzing dead-ends in \textbf{mathematical reasoning} is less straightforward,
because unlike mazes, the valid action space is unconstrained. To simulate an
analogue to backtracking, we utilize a partial mathematical reasoning trajectory
(a non-leaf node from a \btree) and append a ``distractor''---a spurious
reasoning step sampled from a different tree---intended to derail generation
(see Fig. \ref{fig:distractor}). We then evaluate whether the model can still
arrive at the correct final answer, which measures the model's ability to
recover from missteps in reasoning. 

We find that the RL-trained model has significantly improved ability to recover
from distractors compared to their distilled counterpart, with $5.52\%
(\pm 1.76)$, $7.36\% (\pm 1.77)$, $4.18\% (\pm 1.81)$, and $29.09\% (\pm 2.58)$
increase for \texttt{Qwen3-8B}, \texttt{Qwen3.5-9B}, \texttt{Gemma3-12B}, and
\texttt{Qwen2.5-32B}, respectively (see Table \ref{tab:mathbacktrack}). Since
trajectories that do not involve recovery are verifier-invalid, these results
also suggest that the RLVR policy better delineates between valid and invalid
continuations.

\section{Conclusions}
\label{sec:conclusion}

We investigated how RLVR changes test-time exploration through controlled
maze-solving experiments and \btree{}s. Our analysis shows that RLVR-trained
models develop stronger preferences among semantically distinct continuations,
not merely stylistic variants. Sampling experiments on matched
distillation$\rightarrow$RLVR lineages confirm lower empirical semantic entropy
and fewer observed branches under equal sampling budgets. Yet reduced diversity
does not imply poorer recovery, and RLVR improves constraint adherence and
backtracking in mazes, and recovery from distracting steps in mathematical
reasoning.

These findings motivate a simple explanation for RLVR's sampling-efficiency
gains. Better adherence to constraints reduces erroneous actions, and improved
recovery helps trajectories return to valid reasoning; both of these can
increase success rate (Pass@1) without requiring new solution strategies. We
therefore conjecture that improved trace validity drives these gains, while
semantic policy concentration reduces rollout diversity. The resulting challenge
is to preserve semantically diverse trajectories while retaining RLVR's
improvements in validity and recovery.

\section*{Limitations}
\label{sec:limit}

\par\noindent\textbf{Mathematical Reasoning Trees} The \btree construction uses
an LLM-as-a-judge framework with \texttt{GPT-oss-120b} to gauge semantic
equivalence, which might introduce some errors. We performed inter-annotator
agreement tests with more capable frontier LLMs such as \texttt{deepseek-v4-pro}
and OpenAI's \texttt{gpt-5.6-luna}, finding near-perfect agreement (see Appendix
\ref{sec:appendix-aimetree}). We also performed a two-annotator audit of the
judge decisions based on the complete context, and found five errors in 100
samples. We could not perform a large-scale independent expert evaluation owing
to resource constraints.
\par\noindent\textbf{Domain Specificity} The presented empirical evidence is
limited to deterministic environments (mazes) and mathematical reasoning (AIME).
Since we study exploration, this involves sampling a large number of
continuations from frontier LLMs which is prohibitively expensive for some
domains (USD 3000 for collecting AIME traces). Future work that extends this to
domains such as code generation would be compelling.
\par\noindent\textbf{On-policy Distillation} Our aim is to characterize changes
under standard (off-policy/strong-to-weak) distillation and RLVR pipelines,
rather than isolate reward verification from on-policy optimization; on-policy
distillation would address that complementary causal question.

\section*{Ethics Statement}

In keeping with ACL ethical guidelines, all scientific artifacts generated for
this study---including code, prompts, data, and raw model outputs---are made
freely available as open source under the MIT license. Only public datasets
available on the \href{https://huggingface.co/}{Huggingface} platform were used
in the study. AI assistants were not used in ideation or coding, and their
usage was limited to writing tasks (e.g., condensing, checking grammar or tone).
\emph{We do not foresee any potential ethical, societal, or environmental risks
from this work.}

\bibliography{backmatter/references.bib, backmatter/model_refs.bib}
\appendix
\section{\btree Construction}
\label{sec:appendix-aimetree}

To construct \btree{}s, we start with $304$ questions from 
\href{https://huggingface.co/datasets/gneubig/aime-1983-2024}{\textbf{AIME}} as
in \citet{saha2025kismath} and sample responses from a diverse set of LLMs,
namely OpenAI \texttt{o3-2025-04-16}, OpenAI \texttt{o4-mini-2025-04-16}
\citep{gpto}, DeepSeek \texttt{r1-0528} \citep{deepseekr1}, and Qwen's
\texttt{qwen3-235b-a22b-07-25} \citep{qwen3}, with $T=1.0$ and reasoning effort
set to the highest available setting. The number of samples was decided by the
API budget, which in this case totaled \texttt{USD 3000}. All responses with
incorrect answers were removed and this resulted in $\sim20$K responses to $235$
questions, with an average of $\sim85$ responses per question. These were then
split into reasoning segments using OpenAI \texttt{o4-mini-2025-04-16}.

Given a question $Q$, a set of ``correct'' reasoning traces
$A^{(1)}, A^{(2)}, \ldots, A^{(n)}$, and their corresponding segmentations
$a^{(i)}_j$, such that
$A^{(i)} = \textrm{concat}([a^{(i)}_1, a^{(i)}_2, \ldots, a^{(i)}_{n_i}])$, we
employ Algorithm \ref{algo:aimetree} to create \btree{}s. 

\begin{algorithm}
\caption{\btree Construction}
\label{algo:aimetree}
\begin{algorithmic}[1]
    \State Given a question $Q$. 
    \State Given $A^{(i)}~=~\textrm{concat}([a^{(i)}_1, a^{(i)}_2, \ldots, a^{(i)}_{n_i}])$ for $i \in \{1, \ldots n\}$.
    \State Given $M(x, y)$ an LLM-based matching function, \texttt{true} if $y$ is equivalent to $x$, or equivalent to a part of $x$ (subsumed in $x$).
    \State Initialize $T$ with a single root node $v_{\texttt{root}}$ 
    \For{$i \in 1 \ldots n$}
        \State $z[i] \leftarrow v_{\texttt{root}}$
    \EndFor
    \item[]
    \While{$\exists A^{(i)}$ with remaining segments}
        \Statex \textbf{\# Try to consume a segment into a node.}
        \For{ Each $A^{(i)}$ with remaining segments}
            \State $s_{\textrm{head}} \leftarrow \textrm{dequeue}(A^{(i)}); v \leftarrow z[i]$
            \If{ $M(v, s_{\textrm{head}})$ is \texttt{true}}
                \State \textbf{Consume} $s_{\textrm{head}}$ in $v$
            \Else
                \State $\textrm{requeue}(s_{\textrm{head}}, A^{(i)})$
            \EndIf
        \EndFor
        \item[]
        \Statex \textbf{\# Move trace to child node or fork.}
        \For{ Each $A^{(i)}$ with remaining segments}
            \State $s_{\textrm{head}} \leftarrow \textrm{dequeue}(A^{(i)}); v \leftarrow z[i]$
            \State $\texttt{matched} \leftarrow \texttt{false}$
            \For{Each child $v_{\textrm{child}}$ of $v$ in $T$}
                \If{$M(v_{\textrm{child}}, s_{\textrm{head}})$ is \texttt{true}}
                    \State $z[i] \leftarrow v_{\textrm{child}}$
                    \State \textbf{Consume} $s_{\textrm{head}}$ in $v_{\textrm{child}}$
                    \State $\texttt{matched} \leftarrow \texttt{true}$
                    \State \textbf{break}
                \EndIf
            \EndFor
            \If{\texttt{matched} is \texttt{false}}
                \State Create \textbf{new node} $v_{\text{new}}$
                \State Add edge $v \leadsto v_{\text{new}}$
                \State $z[i] \leftarrow v_{\textrm{new}}$
                \State \textbf{Consume} $s_{\textrm{head}}$ in $v_{\textrm{new}}$
            \EndIf
        \EndFor
    \EndWhile
    \State \texttt{return} $T$
\end{algorithmic}
\end{algorithm}

Algorithm \ref{algo:aimetree} starts with a root node, and each trace is
assigned to this root node ($z[i] \leftarrow v_{\texttt{root}}$). At each
iteration, for every trace we try to consume its first available segment into
its assigned node, failing which we either try to match it with one of its
children or spawn a new branch. If the head of the trace matches any of the
children, we consume it at the child, and assign the rest of the trace to the
child. If no such match is found, a new child is spawned, consuming the head
segment, and the rest of the trace is assigned here. The algorithm terminates
when no trace has any unassigned segments. The resulting tree (\btree) has the
property that the lowest common ancestor of two traces represents their longest
shared equivalent reasoning prefix.

The matching function $M(x, y)$ is instantiated with OpenAI \texttt{GPT-oss-120b}
\citep{gptoss} and checks if $x$ is mathematically equivalent to $y$ or $y$ is
mathematically equivalent to a part of $x$. This helps align traces with
differing levels of granularity. In practice, when consuming a segment into a
node, we pick a representative of the node to match against the segment. The LLM
is sampled at $T=1$, $\texttt{top}_p = 0.99$, $\texttt{reasoning\_effort = medium}$,
using 4-shot prompts. A total of $~470$K LLM calls are made for the construction
of the \btree{}s ($\sim2$K calls per \btree). We also measured the
inter-annotator agreement between \texttt{GPT-oss-120b} and more capable
frontier LLMs, namely \texttt{deepseek-v4-pro} \citep{deepseekv4} and
\texttt{gpt-5.6-luna} \citep{gpt56}, and the Cohen's $\kappa$ scores are
$0.815$ and $0.837$, respectively, suggesting that there is near-perfect
agreement.

\begin{figure*}
    \begin{center}
        \includegraphics[width=\textwidth]{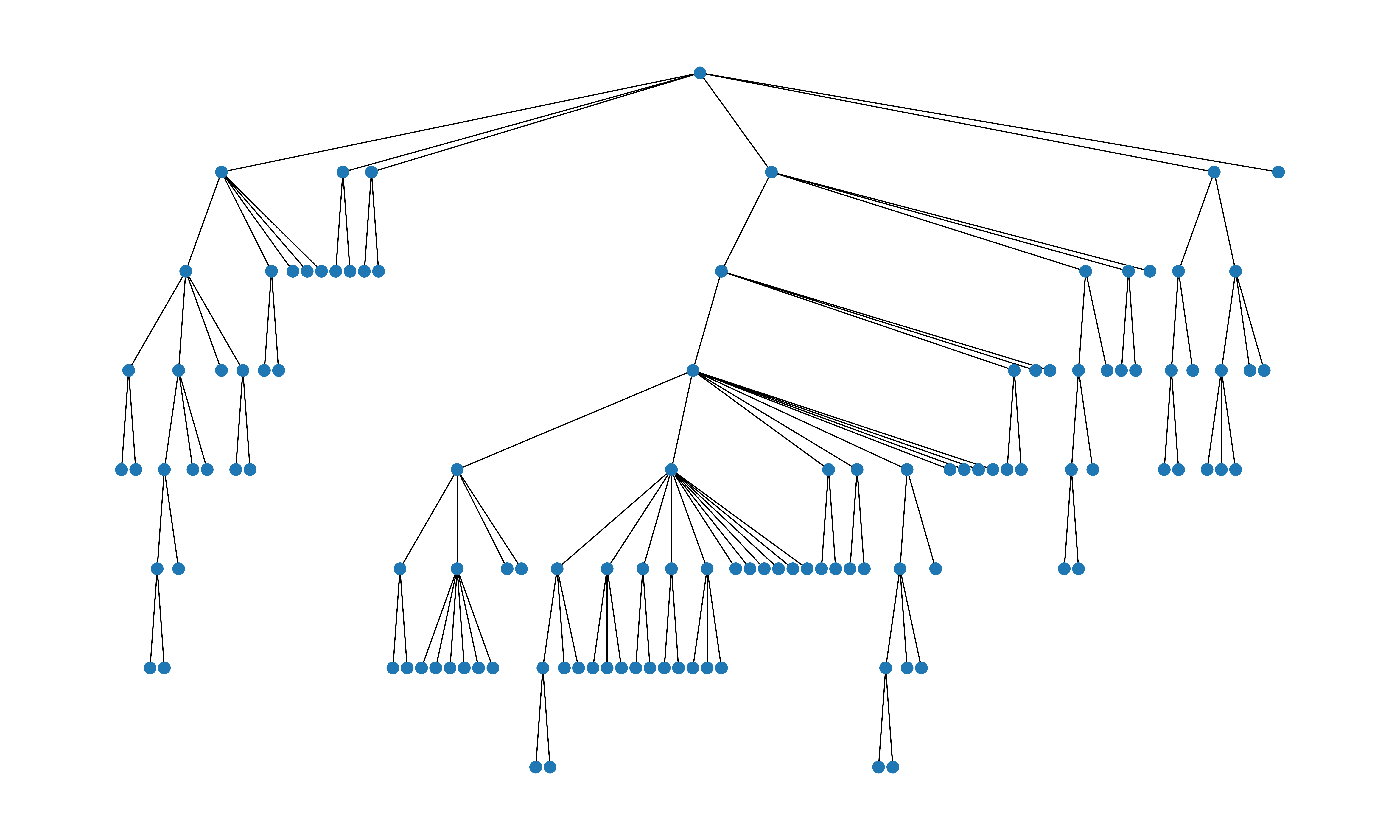}
        \includegraphics[width=\textwidth]{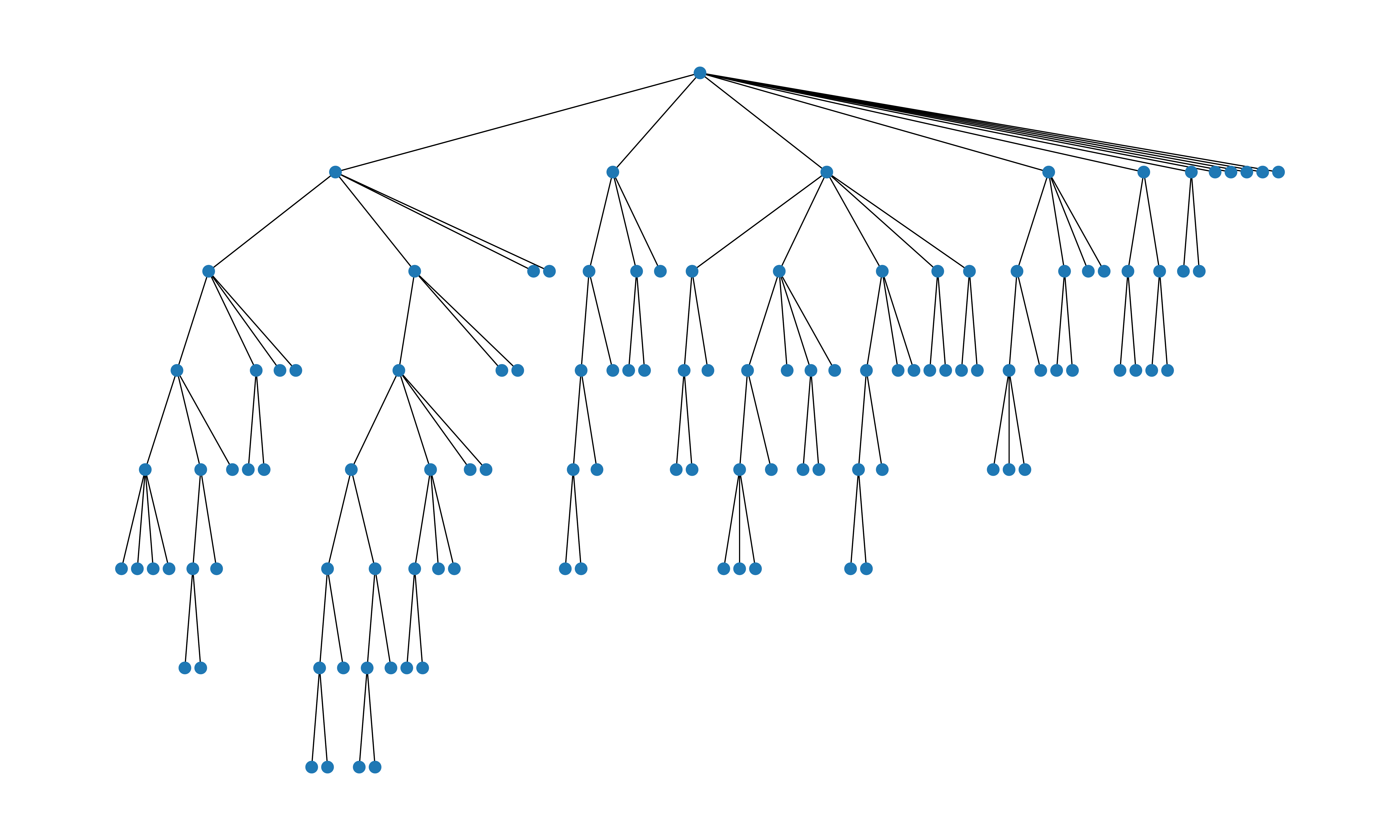}
    \end{center}
    \caption{Examples of extracted \btree{}s.}
    \label{fig:btrees}
\end{figure*}

Further, we performed an audit of $100$ sampled judge decisions with two authors
annotating each sample ($\kappa = 0.90$) with the full context, finding
\textbf{five} errors. Three of these samples had spurious extra text, and two
were identical in content but varied in the level of detail (e.g., skipping some
algebra). The judge is generally robust to semantic perturbations (e.g.,
correctly identified that $\tfrac{u}{\log_b u}$ and $\log_u b^u$ are
equivalent). Crucially, one case had high syntactic overlap but distinct
strategies, which the judge correctly classified. The prompt is given in Fig.
\ref{fig:match-prompt}, some example trees are given in Fig. \ref{fig:btrees},
and some sampled judge decisions are given in Fig. \ref{fig:judgedec}.

\begin{figure*}
    \begin{center}
    \begin{tcolorbox}[width=\textwidth,colback=boxcol,colframe=black]
        \footnotesize{
            \textbf{<system>}\\
            \# Identity

            You are an expert mathematician who gauges similarity of mathematical statements. Help the user detect similarity in statements by answering yes/no.
            \\

            \# Instructions

            \par\noindent* ONLY answer with 'yes' or 'no'. Do not produce any extra text.
            \par\noindent* The user will provide two statements Statement A and Statement B.
            \par\noindent* Answer 'yes' if and only if Statement B is completely equivalent to Statement A or Statement B is equivalent to a part of Statement A.
            \par\noindent* Answer 'no' if the statements are not equivalent, if Statement A is contained in Statement B, or if they have different mathematical meaning.
            \par\noindent* Ignore superficial elements like names of variables, whitespace, synonyms, phrasing, etc.
            \par\noindent* The only important aspect is the mathematical meaning of the statements.
            \\

            \# Examples

                ...

            \textbf{</system>}\\
            \textbf{<user>}\\
            \# Statement A:\\
            The equation of the circle is given by $x^2 + y^2 = 5$.\\
            \# Statement B:\\
            The circle's equation is $z^2 + y^2 = 5$.\\
            \textbf{</user>}\\
            \textbf{<assistant>}\\
            \texttt{yes}\\
            \textbf{</assistant>}
        }
    \end{tcolorbox}
    \end{center}
    \caption{System prompt to instantiate the $M(x, y)$ function in Algorithm \ref{algo:aimetree}.}
    \label{fig:match-prompt}
\end{figure*}

\begin{figure*}
    \begin{center}
    \begin{tcolorbox}[width=\textwidth,colback=boxcol,colframe=black]
        \footnotesize{
            \textbf{Continuation 1:} Convert fractions to common denominator  $4/15=4\times 7=28/105$  $12/35=12\times 3=36/105$  Add them up: $28+36=64/105$.\\
            \textbf{Continuation 2:} Find common denominator: $15*7=105$  $((4*7)/(15*7)) + ((12*3)/(35*3)) = ((28/105)+ (36/105))=64/105$ \\
            \textbf{Judge Decision: \textcolor{red}{Identical}.} 
            \tcbline
            \textbf{Continuation 1:} When $a$ is even, $a/2$ counts for a unique product; when $a$ is odd, $(a-1)/2$ and $(a+1)/2$ give the same product.\\
            \textbf{Continuation 2:} When $m=0,\lfloor a/2\rfloor=a$ the product is the square $m^2$.\\
            \textbf{Judge Decision: \textcolor{red}{Not Identical}.} 
            \tcbline
            \textbf{Continuation 1:} From the second equation we have that $\tfrac{\log_b x}{\log_b(\log_b x)}=54$ or $\tfrac{(u)}{\log_b u}=54$. \\
            \textbf{Continuation 2:} Since $\log_{\log_b x}(x)=54$, $\log_{u} b^{u}=54$\\
            \textbf{Judge Decision: \textcolor{red}{Identical}.} 
            \tcbline
            \textbf{Continuation 1:} So $T$ (or $r$) is formed from the four integers $r_{0}=1,\;r_{1}=3,\;r_{2}=9,\;r_{3}=11,\; r_{4}=101$ and each such $r$ gives rise to a distinct numerator.\\
            \textbf{Continuation 2:} So $\dfrac{T}{9999} = \dfrac{r}{(9999/d)}$, so numerator equals $r$ and $r$ relatively prime to $d$ ($d = 9999/d \cdot (T/r) = 9999/d \cdot d/r = 9999/r$ ($d/r$ is an integer because $\gcd(r,9999/r)=1$).\\
            \textbf{Judge Decision: \textcolor{red}{Not Identical}.} 
        }
    \end{tcolorbox}
    \end{center}
    \caption{Example \emph{LLM-as-a-judge} semantic clustering decisions.}
    \label{fig:judgedec}
\end{figure*}

\section{Experimental Details}
\label{sec:appendix-expdet}

\subsection{Training}

For \textbf{math}, we performed Long-CoT distillation employing $50$K randomly
chosen samples from the
\href{https://huggingface.co/datasets/open-r1/OpenThoughts-114k-math}{\texttt{OpenThoughts-114k-math}}
dataset. We perform Q-LoRA ($r=64, \alpha = 128$; \citet{qlora}) fine-tuning for
$1$ epoch (\texttt{lr} $= 5e-4$) with \texttt{max\_tokens} $= 16$K, batch size
of $64$, a cosine learning rate scheduler with warm-up ($5\%$), and
\texttt{AdamW} optimizer. RL-training was performed with the \texttt{TRL}
library \citep{vonwerra2020trl} using the \texttt{DAPO-Math-17k} dataset
\citep{yu2026dapo} for $1$K steps with \texttt{lr} $= 1e-6$, group size of $8$,
batch size of $48$, and \texttt{max\_tokens} $= 4$K. Following suggestions by
\citet{yu2026dapo}, we set $\epsilon_{\text{low}} = 0.2, \epsilon_{\text{high}} = 0.28$,
and following suggestions by \citet{drgrpo, yu2026dapo, hu2025openreasonerzero},
we set $\beta = 0$. We employ the Dr.~GRPO loss \citep{drgrpo} to avoid
length-based biases. We use the
\href{https://github.com/huggingface/Math-Verify}{Math-Verify}
library to compute accuracy rewards, and use a format reward which ensures that
\texttt{<think>} tags are present and there is exactly one
\texttt{\textbackslash{}boxed\{\}} containing the answer. Length-based rewards
are not employed. \texttt{Qwen3.5-9B}, like \texttt{DAPO-Qwen2.5-32B}, was
trained with Zero-RL, whereas for \texttt{Qwen3-8B} and \texttt{gemma3-12b-pt} we
perform distillation followed by RLVR. The chat templates for the distilled and
RL models are matched, and are distributed alongside the model. \textbf{All
training datasets are de-duplicated against the probing set}.

For \textbf{mazes}, we performed SFT with oracle-generated traces employing the
\href{https://understanding-search.github.io/maze-dataset/maze_dataset.html}{Maze Dataset}
by \citet{maze-dataset}. We used $30$K mazes with sizes
$5\times5,\: 7\times7,\: 9\times9$ and similar settings as described above
(except \texttt{max\_tokens} $= 2$K). For RL-training, $12$K mazes were used
with similar settings as described above (except group size $16$, and
\texttt{max\_tokens}$ = 1.5$K). We used a custom verifier to check correctness,
and a format reward to ensure that the predicted moves were parsable. %
Five tokens were added to the tokenizer before the distillation stage: they are
\texttt{<left>, <right>, <up>, <down>} (for predicting a path), and
\texttt{<-->} (to denote an edge while express the maze as an adjacency list).

RL-training was performed with $4\times4$ H100 $93$GB GPUs, and all other
experiments used $1\times4$ H100s.

\subsection{Evaluation}

\begin{figure*}[h]
    \begin{center}
        \includegraphics[width=\textwidth]{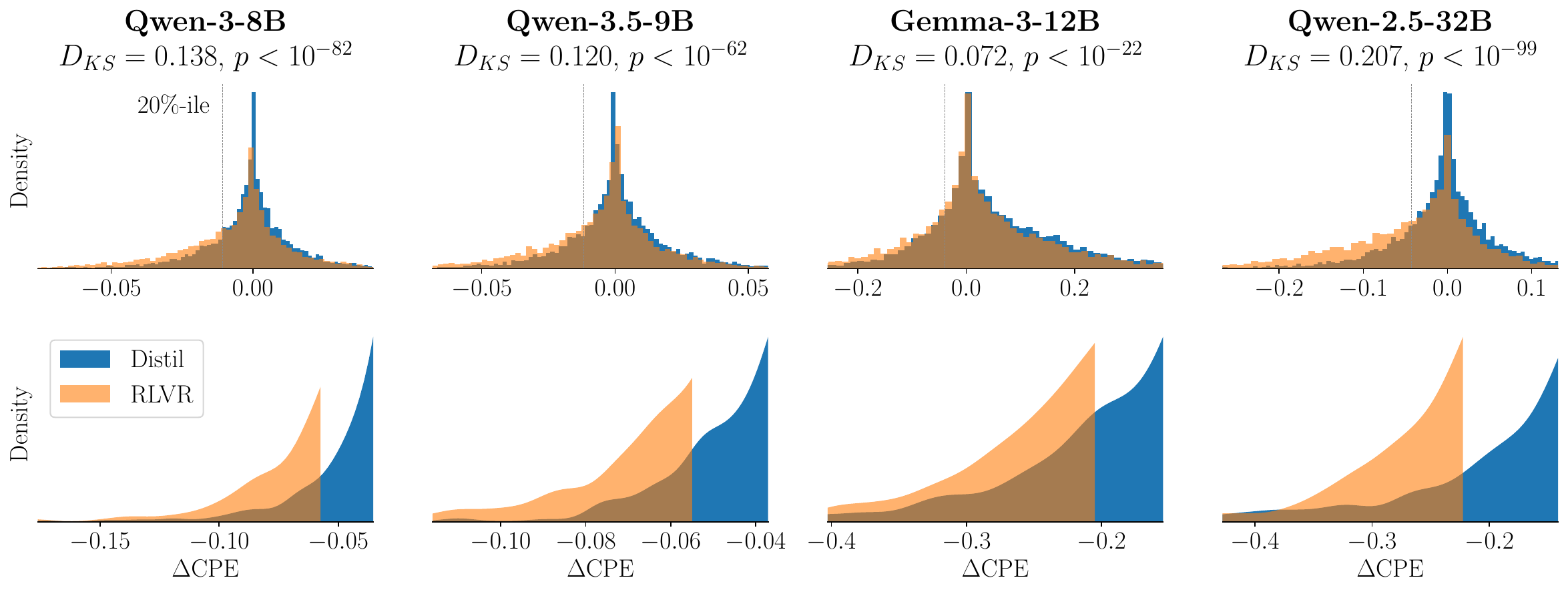}
    \end{center}
    \caption{
        \textbf{Effect of post-training on LLM branching.} (\emph{top}) We plot
        the distribution of \DCPE as described in \S \ref{sec:method} for
        distilled and RLVR-trained models. There is a significant leftward shift
        in the distribution of \DCPE suggesting that RL-trained models have
        enhanced preference at branch points. The (\emph{bottom}) row shows a
        zoomed-in view of the left 20-\%ile tail of each distribution.
    }
    \label{fig:fulldist}
\end{figure*}

\textbf{Unless explicitly mentioned, all sampling parameters are set at default
values,} i.e., $T = 1.0$, and \emph{vanilla random sampling} without
\texttt{top\_k}, \texttt{top\_p}, etc. For results in Fig.~\ref{fig:mazeaccess},
we used $50$ mazes and sampled $1$K generations from the models at $T = 1.0$ and
\texttt{top\_k} $= 10$. For results in \S \ref{subsec:style}, Table
\ref{tab:entdropped}, Fig. \ref{fig:fulldist}, and
Fig. \ref{fig:synt-semant-maze} we directly employed the logits. For \S
\ref{subsec:onpolicy}, we draw $128$ samples at $T = 1.0$ and
\texttt{max\_tokens} $= 512$. The results in \S \ref{subsec:backtrack} are
sampled at $T=1.0$, and an example distractor for the experiment in \S
\ref{subsec:backtrack} is given in Fig. \ref{fig:distractor}.

\begin{figure*}
    \begin{center}
    \begin{tcolorbox}[width=\textwidth,colback=boxcol,colframe=black]
        \footnotesize{
            \textbf{<system>}\\
            You are a helpful AI assistant. Assist the user with their math
            query. The final answer must be in \textbackslash{}boxed\{\}.\\
            \textbf{</system>}\\
            \textbf{<user>}\\
            \textbf{Question}: For non-negative integers $a$ and $b$ with
            $a + b \leq 6$, let $T(a, b) = \binom{6}{a} \binom{6}{b} \binom{6}{a + b}$.
            Let $S$ denote the sum of all $T(a, b)$, where $a$ and $b$ are
            non-negative integers with $a + b \leq 6$. Find the remainder when
            $S$ is divided by $1000$.\\
            \textbf{</user>}\\
            \textbf{<assistant>}\\
            \textcolor{blue}{\texttt{\# Added prefix, BODHI-Tree segments.}}
            \tcbline
            Let $S=\sum_{a=0}^{6}\sum_{b=0}^{6-a}\binom{6}{a}\binom{6}{b}\binom{6}{a+b}$.\\
            Rewrite with $k=a+b$.\\
            $S=\sum_{k=0}^{6}\binom{6}{k}\sum_{a=0}^{k}\binom{6}{a}\binom{6}{k-a}$\\
            $\sum_{a=0}^{k}\binom{6}{a}\binom{6}{k-a}$ is the coefficient of $x^{k}$ in $(1+x)^{6}(1+x)^{6}=(1+x)^{12}$, so it equals $\binom{12}{k}$.\\
            $S=\sum_{k=0}^{6}\binom{6}{k}\binom{12}{k}$\\
            Compute each term:\\
            $k=0:\,1$\\
            $k=1:\,6\cdot12=72$\\
            $k=2:\,15\cdot66=990$\\
            $k=3:\,20\cdot220=4400$\\
            $k=4:\,15\cdot495=7425$\\
            $k=5:\,6\cdot792=4752$
            \tcbline
            \textcolor{red}{Solving gives $c = \frac{4 \pm \sqrt{16 + 240}}{6} = \frac{4 \pm 16}{6}$, so $c = \tfrac{10}{3}$ or $c = -2$.}
            \tcbline
            \textcolor{blue}{\texttt{\# Model continues...}}
            \tcbline
            $k=6:\,\binom{6}{6}\binom{12}{6} = 1 \cdot 924=924 \ldots$
        }
    \end{tcolorbox}
    \end{center}
    \caption{Example distractor (\textcolor{red}{\emph{red}}) for experiment in \S \ref{subsec:backtrack}.}
    \label{fig:distractor}
\end{figure*}

\subsection{Candidate Preference Entropy}

For the experiment in \S \ref{subsec:style}, we note that the branch segments or
candidate continuations from \btree{}s are rather short (see Fig.
\ref{fig:judgedec})---having an average length of $42.96$ tokens
($[42.28, 43.63]$) with the \texttt{Qwen-3} tokenizer, and only $4.51\%$ of the
$7500$ continuations are longer than $100$ tokens. For \CPE calculations in \S
\ref{subsec:style}, we truncate the candidate continuations to $20$ tokens (see
Eq. \ref{eq:bentropy}), this represents a mean coverage of $59.61\%$. The
sequences have distinct surface realisations (average Jaccard similarity for the
sequences truncated to $20$ tokens is $0.2012$ for anchor vs. semantics), and we
ablated this truncation at $40$ and $128$ tokens; it does not change the
conclusions. The \CPE definition averages over the $\operatorname{logprob}$s of
a sequence, to account for length variations. Changing this to a sum of
$\operatorname{logprob}$s  also does not change the conclusions.

\section{Supplementary Results}
\label{sec:appendix-extra}

\begin{table*}
    \begin{center}
        \begin{tabular}[c]{p{8em} p{16em} r}
            \toprule
            Model                       & \textbf{\DCPE Difference}                             & \% with reduced \DCPE         \\
            \midrule
            \texttt{Qwen3-8B}           & $-0.0061$ \:\footnotesize{$[-0.0062, -0.0060]$}       & $95.49$ \:\footnotesize{$[95.07, 95.89]$}      \\
            \texttt{Qwen3.5-9B}         & $-0.0069$ \:\footnotesize{$[-0.0070, -0.0068]$}       & $100.0$ \:\footnotesize{$[100.0, 100.0]$}      \\
            \texttt{Gemma3-12B}         & $-0.0225$ \:\footnotesize{$[-0.0227, -0.0223]$}       & $99.98$ \:\footnotesize{$[99.95, 100.0]$}      \\
            \texttt{Qwen2.5-32B}        & $-0.0345$ \:\footnotesize{$[-0.0349, -0.0340]$}       & $99.46$ \:\footnotesize{$[99.31, 99.60]$}      \\
            \bottomrule
        \end{tabular}
    \end{center}
    \caption{
        \textbf{\btree{}s elucidate policy collapse.} The table shows the change
        in \DCPE
        ($\Delta\textrm{CPE}_{\text{RLVR}} - \Delta\textrm{CPE}_{\text{Distil}}$)
        between distilled and RLVR models, and the percentage of samples where
        \DCPE reduced in the RL-trained model compared to the distilled one.
        There is a significant drop in \DCPE ($p < 0.0001$) for almost all
        samples, suggesting that \emph{RL-trained models have significantly
        stronger trajectory preferences at branch points.}
    }
    \label{tab:entdropped}
\end{table*}

\begin{figure}
    \begin{center}
        \includegraphics[width=\linewidth]{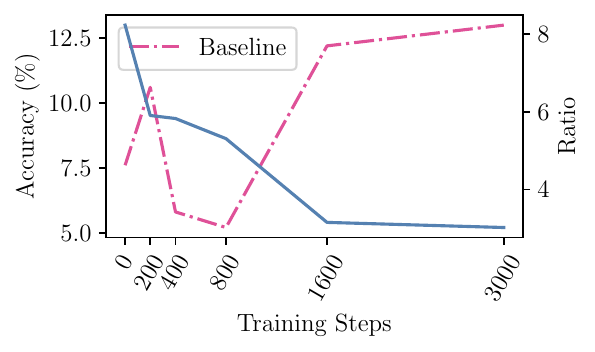}
    \end{center}
    \caption{
        \textbf{Environmental constraint-adherence improves during RL-training.}
        We plot the ratio of the performance with the \texttt{Legal Only}
        intervention and the baseline performance (\emph{dashed line}) of
        various intermediate RL-training checkpoints of \texttt{Qwen3-8B} over
        the course of RL training, which measures the magnitude of the effect
        the \texttt{Legal Only} intervention has. This shows that the policy
        \emph{continually evolves to assign lower probabilities to invalid
        continuations as training progresses}.
    }
    \label{fig:evolution}
\end{figure}

\begin{figure}[h]
    \begin{center}
        \includegraphics[width=\linewidth]{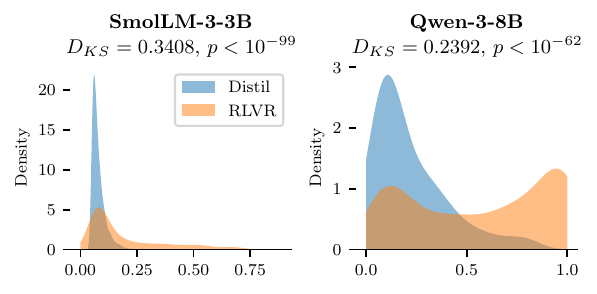}
    \end{center}
    \caption{
        \textbf{RL-trained models are more likely to backtrack.} We plot the
        distribution of probability of backtracking ($p_b$) as described in \S
        \ref{subsec:backtrack}.
    }
    \label{fig:btrack}
\end{figure}

\begin{table*}
    \begin{center}
        \begin{tabular}[c]{l l|c c c r}
            \toprule
            \multirow{2}{*}{\bf Model}  & \multirow{2}{*}{\bf Method}   & \textbf{Original} $(\%)$    & \textbf{Distracted} $(\%)$   & \multirow{2}{*}{\textbf{Unchanged} $(\%)$}    & \multirow{2}{*}{$\Delta (\%)$}       \\
                                        &                               & Acc. [Pass@3]         & Acc. [Pass@3]         &                       &                 \\
            \midrule
            \multirow{2}{*}{\texttt{Qwen3-8B}}        
                                        & Distilled                     & $28.76 \pm 1.85$      & $10.54 \pm 1.26$      & $7.69 \pm 1.09$       & \multirow{2}{*}{$\mathbf{+5.52} \pm 1.76$}  \\
                                        & RLVR                          & $33.61 \pm 1.93$      & $17.73 \pm 1.56$      & $13.21 \pm 1.38$      &   \\
            \midrule
            \multirow{2}{*}{\texttt{Qwen3.5-9B}}      
                                        & Distilled                     & $34.78 \pm 1.95$      & $9.20 \pm 1.18$       & $7.02 \pm 1.04$       & \multirow{2}{*}{$\mathbf{+7.36} \pm 1.77$}  \\
                                        & RLVR                          & $32.11 \pm 1.91$      & $25.08 \pm 1.77$      & $14.38 \pm 1.43$      &   \\
            \midrule
            \multirow{2}{*}{\texttt{Gemma3-12B}}      
                                        & Distilled                     & $23.58 \pm 1.74$      & $11.54 \pm 1.31$      & $9.03 \pm 1.17$       & \multirow{2}{*}{$\mathbf{+4.18} \pm 1.81$}  \\
                                        & RLVR                          & $26.42 \pm 1.80$      & $16.22 \pm 1.51$      & $13.21 \pm 1.38$      &   \\
            \midrule
            \multirow{2}{*}{\texttt{Qwen2.5-32B}}     
                                        & Distilled                     & $70.40 \pm 1.87$      & $60.03 \pm 2.00$      & $52.68 \pm 2.04$      & \multirow{2}{*}{$\mathbf{+29.09} \pm 2.58$}  \\
                                        & RLVR                          & $85.95 \pm 1.42$      & $86.06 \pm 1.38$      & $81.77 \pm 1.58$      &   \\
            \bottomrule
        \end{tabular}
    \end{center}
    \caption{
        \textbf{RLVR models have improved ability to recover from dead-ends.}
        The \textbf{Unchanged} column lists the proportion of samples for which
        the distraction intervention did not make the final answer incorrect,
        and the $\Delta$ column shows the change between the two policies.
    }
    \label{tab:mathbacktrack}
\end{table*}

\end{document}